\documentclass[twoside,11pt]{article}

\usepackage{jmlr2e}
\usepackage{multirow}
\usepackage{color}

\firstpageno{1}

\begin{document}

\title{Causal Learner: A Toolbox for Causal Structure and \\Markov Blanket Learning}

\author{\name Zhaolong Ling \email zlling@ahu.edu.cn \\
       \addr School of Computer Science and Technology, Anhui University, Hefei, Anhui, 230601, China
       \AND
       \name Kui Yu \email yukui@hfut.edu.cn \\
       \addr School of Computer and Information, Hefei University of Technology, Hefei, Anhui, 230009, China
       \AND
       \name Yiwen Zhang \email zhangyiwen@ahu.edu.cn \\
       \addr School of Computer Science and Technology, Anhui University, Hefei, Anhui, 230601, China
       \AND
       \name Lin Liu \email lin.liu@unisa.edu.au \\
       \name Jiuyong Li \email jiuyong.li@unisa.edu.au \\
       \addr  UniSA STEM, University of South Australia, Adelaide, SA, 5095, Australia}

\editor{}

\maketitle

\begin{abstract}

Causal Learner is a toolbox for learning causal structure and Markov blanket (MB) from data.
It integrates functions for generating simulated Bayesian network data, a set of state-of-the-art global causal structure learning algorithms, a set of state-of-the-art local causal structure learning algorithms, a set of state-of-the-art MB learning algorithms, and functions for evaluating algorithms.
The data generation part of Causal Learner is written in R, and the rest of Causal Learner is written in MATLAB.
Causal Learner aims to provide researchers and practitioners with an open-source platform for causal discovery from data and for the development and evaluation of new causal learning algorithms.
The Causal Learner project is available at https://z-dragonl.github.io/causal-learner.
\end{abstract}


\section{Introduction}



Causal networks are graphical models for representing multivariate probability distributions~\citep{pearl2014probabilistic,tsamardinos2006max}. The structure of a causal network takes the form of a directed
acyclic graph (DAG) that captures the causal relationships between variables~\citep{cooper1997simple,spirtes2000causation,koller2009probabilistic}.
Thus, causal structure learning has attracted widespread attention from the machine learning community in recent decades.
Global causal structure learning learns an entire DAG, while local causal structure learning learns only the parents (direct causes) and children (direct effects) of a target variable, as shown in Figure 1 (a) and (b), respectively.
The Markov blanket (MB) in a causal network consists of the parents, children, and spouses (other parents of the target variable's children) of a target variable~\citep{pearl2009causality}, as shown in Figure 1 (c). The MB of a target variable is a minimal set of variables that renders all other variables conditionally independent of the target variable, and thus for a classification problem, the MB of the class attribute is an optimal set for feature selection~\citep{guyon2007causal,aliferis2010local1,aliferis2010local2,yu2020causality}.

\begin{figure}[t]
\centering
 \includegraphics[width=4.3in, height=1.5in]{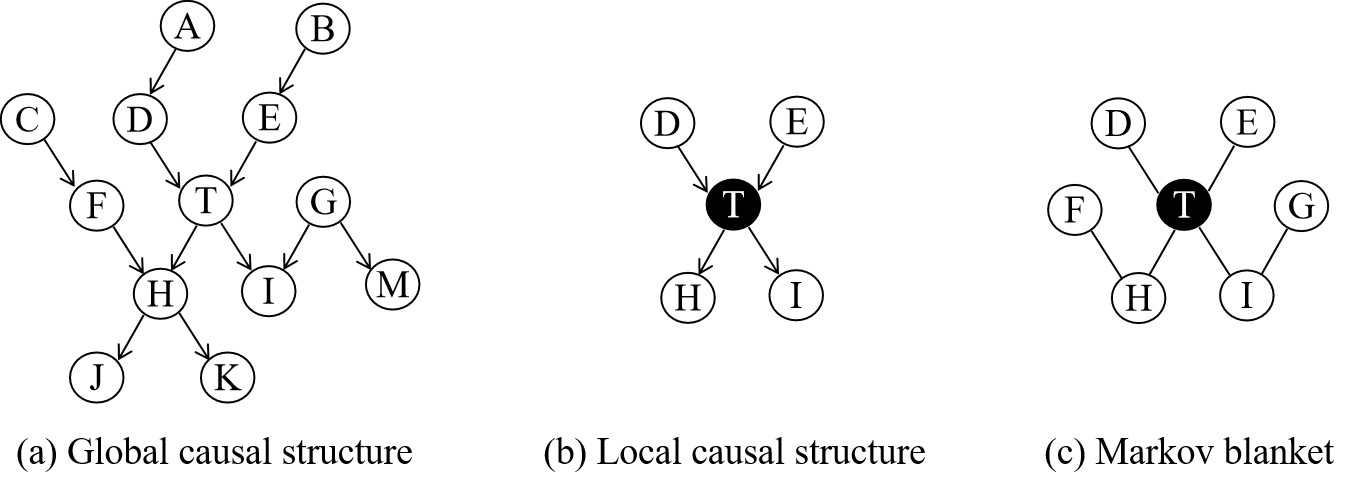}\vspace{-0.4cm}
 \caption{Examples of a global causal structure, local causal structure, and Markov blanket ($T$ in black is a target node).}
 \label{Figure 1}
\end{figure}

To facilitate the research and applications of causal structure learning and MB learning, we develop a toolbox named Causal Learner.
The current well-known toolbox Causal Explorer~\citep{statnikov2010causal} represents the state-of-the-art ten years ago. Compared with Causal Explorer, there are three main contributions of Causal Learner. (1) It offers more state-of-the-art algorithms than Causal Explorer. (2) It offers functions for generating simulated data from Bayesian networks (BNs) and functions for evaluating the performance of algorithms, which are not provided by Causal Explorer. (3) It is completely open-source, which makes it easier for researchers and practitioners to understand, modify, and apply, while the source code of Causal Explorer is not provided.



\begin{figure}[!htbp]
\centering
 \includegraphics[width=4.4in, height=1.55in]{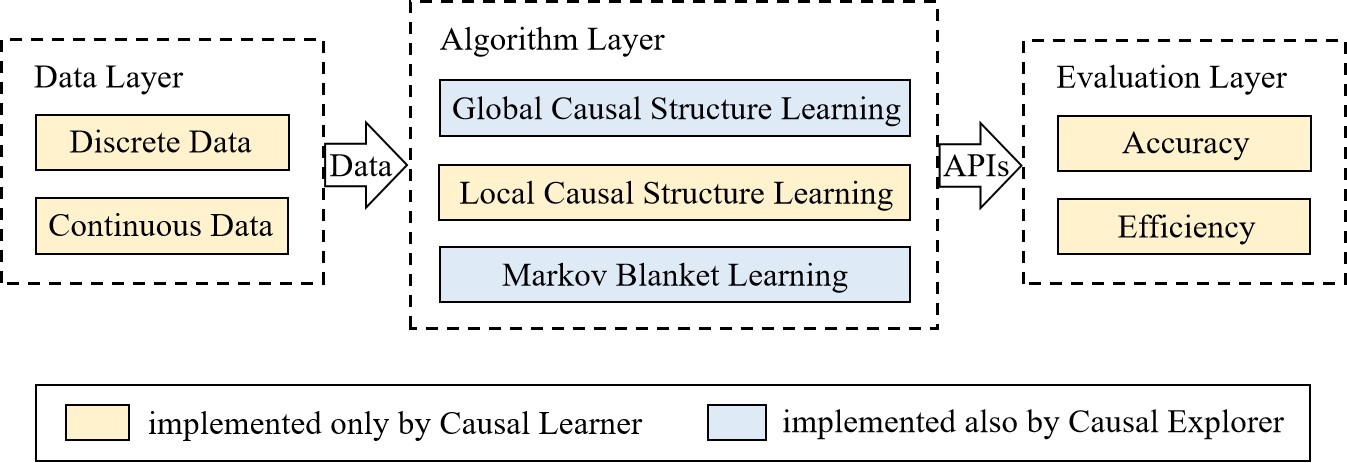}
 \caption{The architecture of Causal Learner.}
 \label{Figure 1}
\end{figure}

\section{Architecture}

Figure 2 shows the hierarchical architecture of Causal Learner, in comparison with Causal Explorer. As Causal Explorer was developed 10 years ago, it does not contain many new algorithms, and it does not have a data generation or evaluation component. By contrast, Causal Learner conceives a more ambitious blueprint. It aims to support the entire causal structure and MB learning procedure, including data generation, state-of-the-art algorithms, and algorithm evaluation.

\begin{table}[t]
\centering
\scriptsize

\begin{tabular}{|c|lcc|lcc|}
\hline

                                                                                         & Name       & \#Nodes & \#Arcs & Name                  & \#Nodes    & \#Arcs     \\\hline
\multirow{10}{*}{\begin{tabular}[c]{@{}c@{}}Discrete\\ Bayesian\\ Network\end{tabular}}  & CANCER     & 5     & 4    & BARLEY                & 48       & 84       \\
                                                                                         & EARTHQUAKE & 5     & 4    & HAILFINDER            & 56       & 66       \\
                                                                                         & SURVEY     & 6     & 6    & HEPAR II              & 70       & 123      \\
                                                                                         & ASIA       & 8     & 8    & WIN95PTS              & 76       & 112      \\
                                                                                         & SACHS      & 11    & 17   & PATHFINDER            & 109      & 195      \\
                                                                                         & CHILD      & 20    & 25   & ANDES                 & 223      & 338      \\
                                                                                         & INSURANCE  & 27    & 52   & DIABETES              & 413      & 602      \\
                                                                                         & WATER      & 32    & 66   & PIGS                  & 441      & 592      \\
                                                                                         & MILDEW     & 35    & 46   & LINK                  & 724      & 1125     \\
                                                                                         & ALARM      & 37    & 46   & MUNIN (4 subnetworks) & 186 -1041 & 273 -1388 \\\hline
\multirow{3}{*}{\begin{tabular}[c]{@{}c@{}}Continuous\\ Bayesian\\ Network\end{tabular}} & SANGIOVESE & 15    & 55   & ECOLI70               & 46       & 70       \\
                                                                                         & MEHRA      & 24    & 71   & MAGIC-IRRI            & 64       & 102      \\
                                                                                         & MAGIC-NIAB & 44    & 66   & ARTH150               & 107      & 150      \\\hline
\end{tabular}
\caption{Benchmark Bayesian networks.}
\end{table}

\subsection{Data}

In the data layer, Causal Learner generates two types of data: discrete data and continuous data.
The data are generated based on various benchmark BNs (written in R), and the details of each BN are shown in Table 1.
The data can also be generated by the bnlearn~\citep{jstatsoft09} toolbox, but the generated data are encapsulated in R language classes and cannot be easily used by researchers using other programming languages. Causal Learner can output the generated data as text for easy and flexible use.

\subsection{Algorithm}

In the algorithm layer, Causal Learner implements 7 global causal structure learning algorithms, 4 local causal structure learning algorithms, and 15 MB learning algorithms (written in MATLAB). Table 2 lists all of these algorithms.
To ensure the correctness of the algorithms implemented in Causal Learner, unless the original implementations of the algorithms are not released, we always try to integrate the original versions rather than re-implement them. Additionally, we have used the same data to evaluate the algorithms in Causal Learner. Compared with Causal Explorer, the results of Causal Learner are comparable in accuracy and much more efficient.




\subsection{Evaluation}

In the evaluation layer, Causal Learner provides abundant metrics for evaluating causal structure
and MB learning algorithms (written in MATLAB), including 10 metrics for evaluating accuracy and 2 metrics for evaluating efficiency.


\begin{table}[t]
\centering
\scriptsize

\begin{tabular}{|l|c|c|l|c|c|l|c|c|}

\hline
\multicolumn{3}{|c|}{\multirow{2}{*}{Global Causal Structure Learning}} & \multicolumn{3}{c|}{\multirow{2}{*}{Local Causal Structure Learning}} & \multicolumn{3}{c|}{\multirow{2}{*}{Markov Blanket Learning}} \\
\multicolumn{3}{|c|}{}                                                  & \multicolumn{3}{c|}{}                                                 & \multicolumn{3}{c|}{}                                         \\\hline
Algorithm     & Learner          & Explorer         & Algorithm      & Learner           & Explorer      & Algorithm    & Learner      & Explorer     \\\hline
SCA                    & $\circ$               & $\bullet$            & PCD-by-PCD             & $\bullet$             & $\circ$             & GS                   & $\bullet$         & $\bullet$         \\
PC                     & $\bullet$             & $\bullet$            & MB-by-MB               & $\bullet$             & $\circ$             & IAMB                 & $\bullet$         & $\bullet$         \\
TPDA                   & $\circ$               & $\bullet$            & CMB                    & $\bullet$             & $\circ$             & interIAMB            & $\bullet$         & $\bullet$         \\
GES                    & $\bullet$             & $\circ$              & LCS-FS                 & $\bullet$             & $\circ$             & IAMBnPC              & $\bullet$         & $\bullet$         \\
GSBN                   & $\bullet$             & $\circ$              &                        &                       &                     & interIAMBnPC         & $\bullet$         & $\bullet$         \\
MMHC                   & $\bullet$             & $\bullet$            &                        &                       &                     & Fast-IAMB            & $\bullet$         & $\bullet$         \\
PC-stable              & $\bullet$             & $\circ$              &                        &                       &                     & FBED                 & $\bullet$         & $\circ$           \\
F2SL-c                 & $\bullet$             & $\circ$              &                        &                       &                     & MMMB                 & $\bullet$         & $\bullet$         \\
F2SL-s                 & $\bullet$             & $\circ$              &                        &                       &                     & HITON-MB             & $\bullet$         & $\bullet$         \\
                       &                       &                      &                        &                       &                     & PCMB                 & $\bullet$         & $\circ$           \\
                       &                       &                      &                        &                       &                     & IPCMB                & $\bullet$         & $\circ$           \\
                       &                       &                      &                        &                       &                     & MBOR                 & $\bullet$         & $\circ$           \\
                       &                       &                      &                        &                       &                     & STMB                 & $\bullet$         & $\circ$           \\
                       &                       &                      &                        &                       &                     & BAMB                 & $\bullet$         & $\circ$           \\
                       &                       &                      &                        &                       &                     & EEMB                 & $\bullet$         & $\circ$      \\\hline
\end{tabular}

\caption{
Algorithms included in (indicated by $\bullet$) and absent from (indicated by $\circ$) Causal Learner and Causal Explorer.
}
\end{table}

\begin{figure}[!htbp]
\centering
 \includegraphics[width=5.8in, height=2.9in]{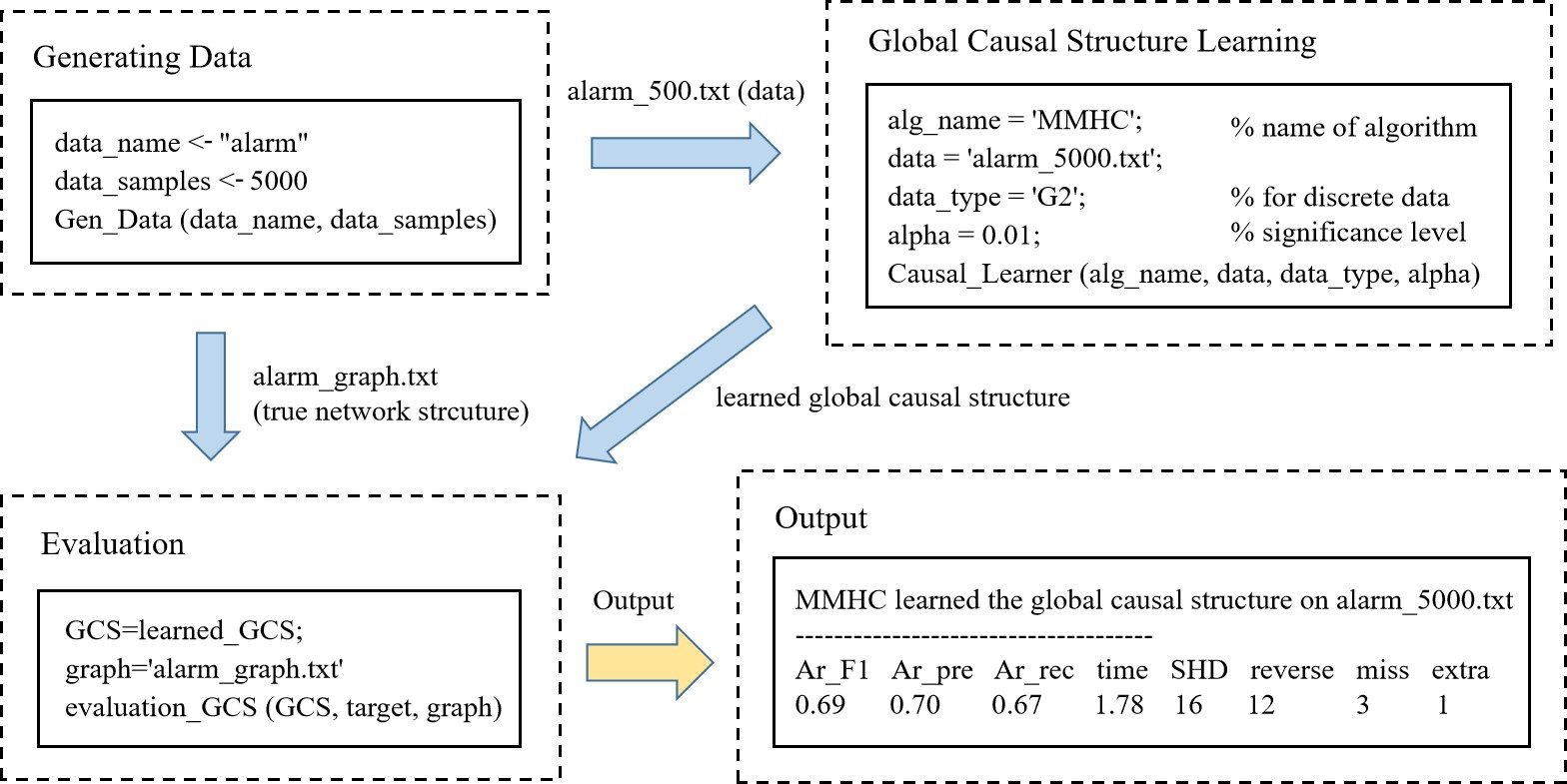}
 \caption{An example of using Causal Learner to learn a global causal structure.}
 \label{Figure 1}
\end{figure}

\section{Usage Example}

Causal Learner comes with a manual that details the BNs, algorithms, evaluation metrics, and how each function is used, https://github.com/z-dragonl/Causal-Learner. Figure 3 shows an example of global causal structure learning using Causal Learner.
As shown in Figure 3, Causal Learner needs only 4 input parameters when learning a global causal structure, while Causal Explorer requires additional parameters such as ``domain\_count". Thus, Causal Learner uses a cleaner input format than that of Causal Explorer.

\section{Conclusion and Future Work}

Causal Learner is an easy-to-use open-source toolbox for causal structure learning and MB learning, which aims to promote research progress in the causal discovery community. The current version of Causal Learner includes simulated BN data generation functions, causal structure and MB learning algorithms, and algorithm evaluation functions. Causal Learner is still growing. Future work includes extending Causal Learner with causal structure and MB learning algorithms without causal sufficiency or faithfulness assumptions.



\acks{We would like to acknowledge support for this project from the National Key Research and Development Program of China (under grant 2020AAA0106100), and the National Science Foundation of China (under
grant U1936220 and 61872002).}


%
%
%
%
%
%
%

\vskip 0.2in
\bibliography{References}

\begin{thebibliography}{12}
\providecommand{\natexlab}[1]{#1}
\providecommand{\url}[1]{\texttt{#1}}
\expandafter\ifx\csname urlstyle\endcsname\relax
  \providecommand{\doi}[1]{doi: #1}\else
  \providecommand{\doi}{doi: \begingroup \urlstyle{rm}\Url}\fi

\bibitem[Aliferis et~al.(2010{\natexlab{a}})Aliferis, Statnikov, Tsamardinos,
  Mani, and Koutsoukos]{aliferis2010local1}
Constantin~F Aliferis, Alexander Statnikov, Ioannis Tsamardinos, Subramani
  Mani, and Xenofon~D Koutsoukos.
\newblock Local causal and markov blanket induction for causal discovery and
  feature selection for classification part i: Algorithms and empirical
  evaluation.
\newblock \emph{Journal of Machine Learning Research}, 11\penalty0
  (Jan):\penalty0 171--234, 2010{\natexlab{a}}.

\bibitem[Aliferis et~al.(2010{\natexlab{b}})Aliferis, Statnikov, Tsamardinos,
  Mani, and Koutsoukos]{aliferis2010local2}
Constantin~F Aliferis, Alexander Statnikov, Ioannis Tsamardinos, Subramani
  Mani, and Xenofon~D Koutsoukos.
\newblock Local causal and markov blanket induction for causal discovery and
  feature selection for classification part ii: Analysis and extensions.
\newblock \emph{Journal of Machine Learning Research}, 11\penalty0
  (Jan):\penalty0 235--284, 2010{\natexlab{b}}.

\bibitem[Cooper(1997)]{cooper1997simple}
Gregory~F Cooper.
\newblock A simple constraint-based algorithm for efficiently mining
  observational databases for causal relationships.
\newblock \emph{Data Mining and Knowledge Discovery}, 1\penalty0 (2):\penalty0
  203--224, 1997.

\bibitem[Guyon et~al.(2007)Guyon, Aliferis, et~al.]{guyon2007causal}
Isabelle Guyon, Constantin Aliferis, et~al.
\newblock Causal feature selection.
\newblock In \emph{Computational methods of feature selection}, pages 75--97.
  Chapman and Hall/CRC, 2007.

\bibitem[Koller et~al.(2009)Koller, Friedman, and
  Bach]{koller2009probabilistic}
Daphne Koller, Nir Friedman, and Francis Bach.
\newblock \emph{Probabilistic graphical models: principles and techniques}.
\newblock MIT press, 2009.

\bibitem[Pearl(2009)]{pearl2009causality}
Judea Pearl.
\newblock \emph{Causality}.
\newblock Cambridge university press, 2009.

\bibitem[Pearl(2014)]{pearl2014probabilistic}
Judea Pearl.
\newblock \emph{Probabilistic reasoning in intelligent systems: networks of
  plausible inference}.
\newblock Elsevier, 2014.

\bibitem[Scutari(2010)]{jstatsoft09}
M.~Scutari.
\newblock {Learning Bayesian Networks with the bnlearn R Package}.
\newblock \emph{Journal of Statistical Software}, 35\penalty0 (3):\penalty0
  1--22, 2010.
\newblock URL \url{http://www.jstatsoft.org/v35/i03/}.

\bibitem[Spirtes et~al.(2000)Spirtes, Glymour, and
  Scheines]{spirtes2000causation}
Peter Spirtes, Clark~N Glymour, and Richard Scheines.
\newblock \emph{Causation, prediction, and search}.
\newblock MIT press, 2000.

\bibitem[Statnikov et~al.(2010)Statnikov, Tsamardinos, Brown, and
  Aliferis]{statnikov2010causal}
Alexander Statnikov, Ioannis Tsamardinos, Laura~E Brown, and Constantin~F
  Aliferis.
\newblock Causal explorer: A matlab library of algorithms for causal discovery
  and variable selection for classification.
\newblock \emph{Causation and Prediction Challenge Challenges in Machine
  Learning, Volume 2}, pages 267--278, 2010.

\bibitem[Tsamardinos et~al.(2006)Tsamardinos, Brown, and
  Aliferis]{tsamardinos2006max}
Ioannis Tsamardinos, Laura~E Brown, and Constantin~F Aliferis.
\newblock The max-min hill-climbing bayesian network structure learning
  algorithm.
\newblock \emph{Machine learning}, 65\penalty0 (1):\penalty0 31--78, 2006.

\bibitem[Yu et~al.(2020)Yu, Guo, Liu, Li, Wang, Ling, and Wu]{yu2020causality}
Kui Yu, Xianjie Guo, Lin Liu, Jiuyong Li, Hao Wang, Zhaolong Ling, and Xindong
  Wu.
\newblock Causality-based feature selection: Methods and evaluations.
\newblock \emph{ACM Computing Surveys (CSUR)}, 53\penalty0 (5):\penalty0 1--36,
  2020.

\end{thebibliography}

\end{document}